\documentclass[10pt,twocolumn,letterpaper]{article}

\usepackage{cvpr}
\usepackage{times}
\usepackage{epsfig}
\usepackage{graphicx}
\usepackage{amsmath}
\usepackage{amssymb}
\usepackage{algorithm}
\usepackage[noend]{algpseudocode}
\usepackage{epstopdf}
\usepackage[toc,page]{appendix}

\newcommand{\Cal}[1]{{\cal #1}}

\cvprfinalcopy 
\usepackage[pagebackref=true,breaklinks=true,letterpaper=true,colorlinks,bookmarks=false]{hyperref}
\begin{document}
	
	\title{The Next Best Underwater View}
	
	\author{Mark Sheinin and Yoav Y. Schechner\\
		Department of Electrical Engineering\\
		Technion - Israel Inst. of Technology, Haifa 32000, Israel\\	
     	{\tt\small markshe@campus.technion.ac.il , yoav@ee.technion.ac.il}		
	}
\maketitle

\begin{abstract}
 To image in high resolution large and occlusion-prone scenes, a camera must move
 above and around. Degradation of visibility due to geometric occlusions and distances 
 is exacerbated by scattering, when the scene is in a participating medium. Moreover, underwater
 and in other media, artificial lighting is needed. Overall, data quality depends on the observed surface, medium and the time-varying poses of the camera and light source (C\&L). This work proposes to optimize C\&L poses as they move, so that
 the surface is scanned efficiently and the descattered recovery has the highest quality. The work generalizes the
 {\em next best view} concept of robot vision to scattering media and cooperative movable lighting. It also extends
 descattering to platforms that move optimally. The optimization criterion is {\em information gain}, taken from information theory. We exploit the existence of a prior rough 3D model, since
 underwater such a model is routinely obtained using sonar. We demonstrate this principle in a scaled-down setup.
\end{abstract}


\section{Introduction}
Scattering media degrades images. Studies aimed at enhancing visibility
focus on single-image dehazing~\cite{fattal2008single,he2011single}, or methods that modulate
properties of the illumination, such as spatio-temporal
structure~\cite{dalgleish2013extended,gkioulekas2015micron,gupta2008controlling,o2012primal,heide2014imaging}, polarization~\cite{treibitz2009active,treibitz2012turbid}.
However, all these methods do not exploit an important degree of freedom: the dynamic
pose of the camera.

Pose dynamics is important, because most imaging platforms move {\em anyway}. Even without a participating
medium, a camera must move around to view large areas and zones behind objects and concavities~\cite{anguelov2010google}.
Platform motion, however, needs to be efficient, covering the surface domain in the highest
quality, in the shortest time. The camera needs to move, so that object regions that have not been well observed, will be efficiently recovered next. This is the {\em next best view} (NBV) concept in robot vision.
Prior NBV designes assumed no participating medium, being ruled solely
by object occlusions. However, a scattering medium, not only occlusions, disrupt visibility.
This affects drones overflying wide hazy scenes, autonomous underwater vehicles that scan the sea floor
and inspect submerged infrastructure and fire-fighting rovers that operate in smoke. Despite their motion and need to overcome scatter, existing systems scan scenes~\cite{campos2014surface} while ignoring scattering.
\begin{figure}[t]
	\def\svgwidth{\columnwidth}
	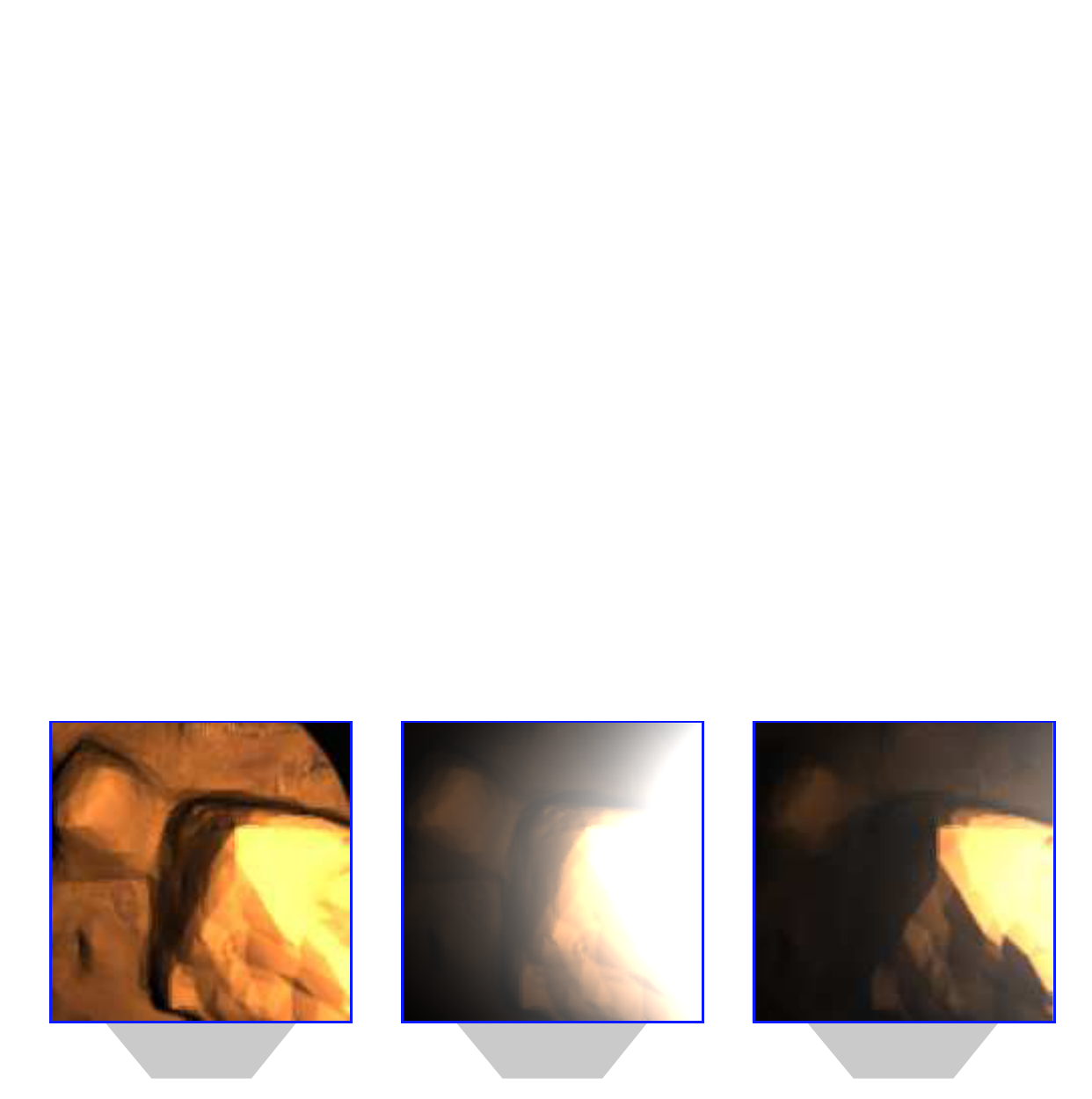
	\caption{\small Underwater Imaging trade-offs. (a) A camera (red) is pointed at the surface. Two possible positions for lighting the scene are visible. The lighting positions differ in the degree of separation from the camera. (b) Image created from the a small separation distance (yellow) in non scattering medium. (c) Image created from the same - small separation distance but in a scattering medium. Backscatter significantly reduces image quality. (d) Light is positioned farther from the camera (blue). Backscatter is reduced but surface details are lost due to shadowing effects.}
	\label{fig:UnderwaterTradeOff}
\end{figure}

This work generalizes NBV to scattering media. We achieve 3D descattering in large areas
and around occlusions, through sequential changes of pose. The obvious need to move the platform
in large areas and occlusions is {\em exploited} for optimized dehazing, i.e, estimation of surface albedo.
On the other hand, scattering by the medium influences the optimal changes of pose. The challenge is exacerbated when lighting must be brought-in, in deep underwater operations, tissue and indoor smoky scenes. Scattering affects object irradiance and volumetric backscatter~\cite{gupta2008controlling,jaffe1990computer}, as a function of the {\em lighting pose}, not only the {\em camera pose} (Fig.~\ref{fig:UnderwaterTradeOff}). Usually both the camera and lighting ({\tt C}\&{\tt L}) are mounted on the same rig. However, visibility can potentially be enhanced using separate platforms~\cite{jaffe2007multi}. Therefore, the {\em next best underwater view} (NBUV) optimizes the next joint poses of {\tt C}\&{\tt L}.
 
The optimization criterion is {\em information gain}, taken from information theory. We exploit the existence of a prior rough 3D model, since underwater such a model is routinely obtained using active sonar. We demonstrate this principle in scaled-down experimentation.
\begin{figure}[t]
	\def\svgwidth{\columnwidth}
	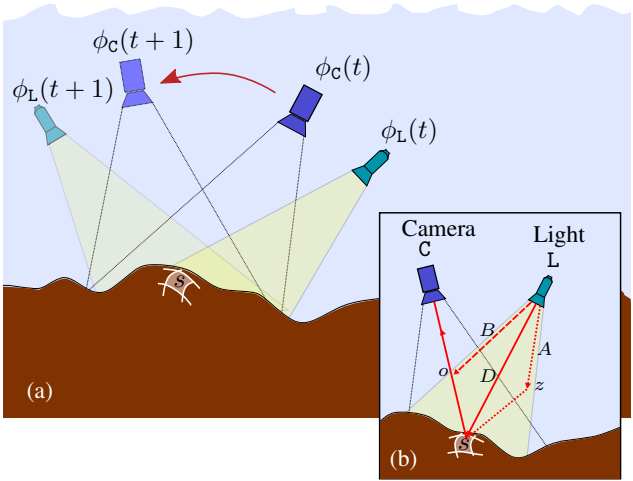
	\caption{\small (a) The next best underwater view task seeks camera and light poses ${\bf \phi}_{\tt C}(t+1)$, ${\bf \phi}_{\tt L}(t+1)$ that maximize the information gain. (b) The sensed radiance of surface patch $ s $ is comprised of 3 illumination signals: the direct ($D_s$) and ambient ($A_s$) illumination signals, and parasitic backscatter ($B_s$).}
	\label{fig:ProblemFormulation}
\end{figure}

\section{Theoretical Background}
\label{sec:background}

\subsection{Imaging in a medium}
\label{sec:underwaterim}
Consider Fig.~\ref{fig:ProblemFormulation}b. At time $ t $, the pose of light source {\tt L} has a vector of location and orientation parameters, ${\bf \phi}_{\tt L}(t)$.
The source irradiates submerged surface patch $s$ from distance $l_\text{LS}$. 
The medium has extinction coefficient $\beta$. 
The surface irradiance~\cite{gupta2008controlling} at $s$ is
\begin{eqnarray}
\tilde E_{s} = D_{s} + A_{s}
\label{eq:E}
\end{eqnarray}
The component $D_{s}$ is due to direct transmission from {\tt L} to $s$, while $A_{s}$
is due to ambient indirect surface illumination. The latter is mainly created by off-axis scattering of the illumination beam. The surface illumination decreases exponentially with $ l_\text{LS} $:
\begin{eqnarray}
\label{eq:Edecay}
&D_s \propto \Cal{C}_0 \exp(-\beta l_\text{LS})/ l_\text{LS}^2\\
&A_s \propto \Cal{C}_0 \exp(-\beta \left[l_\text{Lz}^2 + l_\text{Sz}^2\right])/ l_\text{Lz}^2 / l_\text{Sz}^2,
\end{eqnarray}
where $ \Cal{C}_0 $ is the intensity of {\tt L}.
The object signal is  $\rho_s E_{s}$, where $\rho_s$ is the albedo at $s$ and
\begin{equation}
  E_{s} = \tilde E_{s} \exp(-\beta l_\text{SC}).
  \label{eq:E2}
\end{equation}
Here $l_\text{SC}$ is the distance from $s$ to camera {\tt C}. At time $t$, the pose of {\tt C} is represented by a vector of parameters ${\bf \phi}_{\tt C}(t)$.
The line of sight from {\tt C} to patch $s$ includes backscatter $B_{s}$, which increases~\cite{jaffe1990computer,treibitz2012turbid} with $l_\text{So}$. 

The imaged radiance~\cite{cowan1992automatic} is 
\begin{equation}
  I_s = \rho_s E_s + B_s + n_I, 
\label{eq:ImageCreationModel}
\end{equation}
where $n_I$ is noise. Note that in this paper, all the radiometric terms ($ I_s,~E_s,~n_I,$ etc..) are in photoelectron units [e].  The noise $n_I$ has two components~\cite{ratner2007illumination}: photon noise and the read noise. The variance of photon noise is $\sigma_{\text{PN}}^2 = I_s$. Readout noise is assumed to be signal-independent, with variance $ \sigma^2_{\text{RN}} $.
The probability distribution function (PDF) of $ n_I $ is approximately Gaussian with variance:
\begin{equation}
\label{eq:sigmaI}
 \sigma_I^2 =  \sigma_{\text{PN}}^2 + \sigma_{\text{RN}}^2 .
\end{equation}

The patches's signal-to-noise ratio (SNR) is therefore:
\begin{equation}
\text{SNR}_s \approx \frac{I_{s}}{ \sqrt{ \sigma_{\text{PN}}^2 + \sigma_{\text{RN}}^2} } \approx \frac{ \rho_s E_{s} }{\sqrt{\rho_s E_{s} + B_{s} +\sigma_{\text{RN}}^2}}.
\label{eq:SNRs}
\end{equation}
In a clear medium, backscatter is negligible $ B_s\ll\sigma_{RN}^2 $. Under sufficient lighting: $ \text{SNR}_s \sim \sqrt{\rho_sE_{s}}$ (See~\ref{eq:SNRs}). Thus,  at the best SNR, $E_s$ is maximized. This is achieved by avoiding shadows~\cite{sakane1991automatic}, i.e., placing {\tt L} very close to {\tt C} (Fig.~\ref{fig:UnderwaterTradeOff}).

Underwater, placing {\tt L} very close to {\tt C} results in significant backscatter $ B_s $, which reduces $ \text{SNR}_s $ in (\ref{eq:SNRs}). To reduce backscatter, {\tt L} is usually separated from $ \tt C $. Such a separation may result in shadows. In a shadow, $E_s\ll\sigma_{\text{RN}}^2$ 
while the extinction of light (Eqs.~\ref{eq:Edecay}-\ref{eq:E2}) compounds this effect. Thus optimal setting of {\tt L} underwater is non-trivial.

At time $ t $, the projection of $s$ to the camera  frame is:
\begin{equation}
 \label{eq:CamProjection}
 {\cal P}_t~:~s\rightarrow U_s(t),
\end{equation}
where $ U_s(t) $ is a set of image pixels $ \bf x $ that patch $ s $ is projected to.
Therefore, in the camera frame, the components of Eqs.~(\ref{eq:E2},\ref{eq:ImageCreationModel})  are \mbox{$E({\bf x},t) = {\cal P}(E_s)$}, $B({\bf x},t) = {\cal P}(B_s)$  and 
$I({\bf x},t) = {\cal P}(I_s)$.


\subsection{Next Best View}
\label{sec:nbv}

The NBV task is generally formulated as follows. Let $\cal O$ represent a property of the object, e.g., the  spatially varying albedo or topography. A computer vision system estimates this representation, $\hat{\cal O}$, using sequential measurements. By $t$, the camera has already accumulated image data $I(t')$,  ~$~\forall t'\leq t$. All this preceding data is processed to yield $\hat{\cal O}(t)$. Let $\Phi_{\tt C}$ be the set of all possible camera poses. A {\em next view} is planned for time $t+1$, where the camera may be posed at ${\bf \phi}_{\tt C}(t+1)\in \Phi_{\tt C}$, yielding new data. The new data helps getting an improved estimate
$\hat{\cal O}(t+1)$.
The NBV question is: out of all possible views in $\Phi_{\tt C}$, what is the {\em best} ${\bf \phi}_{\tt C}(t+1)$, such that $\hat{\cal O}(t+1)$ has the best quality? Formulating this task mathematically depends on a quality criterion, prior knowledge about $\cal O$, and the type of camera; e.g., passive or active 3D scanner. Different studies have looked at different aspects of the NBV task \cite{chen2005vision,pito1999solution,wenhardt2006information}. Nevertheless, they were all designed for imaging in clear media.


\subsection{Information Gain}
\label{sec:infgain}

Consider a random variable $a$. Let $f(a)$ be its PDF. The differential entropy \cite{ahmed1989entropy} of $a$ is then
\begin{equation}
	H(a)=-\int f(a) \ln [f(a)]da.
	\label{eq:Hdef}
\end{equation}
At time $t$, the variable $a$ has entropy $H_t(a)$. Then, at time $t+1$, new data decreases the uncertainty of $a$, consequently the PDF of $ a $ is narrowed and its differential entropy decreases $H_{t+1}(a)<H_t(a)$. The {\em information gain} due to the new data is then \cite{norwich1993information} defined by
\begin{equation}
	{\cal I}_{t+1}(a) = H_t(a)- H_{t+1}(a).
	\label{eq:Idef}
\end{equation}
Suppose $a$ is normally distributed, with variance $\sigma_a^2(t)$ and $\sigma_a^2(t+1)$
at $t$ and $t+1$ respectively. Then Eqs.~({\ref{eq:Hdef},\ref{eq:Idef}) yield
	
\begin{equation}
	H_t(a)=(1/2) \ln[2\pi e \sigma_a^2(t)],
	\label{eq:Hgauss}
\end{equation}
\begin{equation}
	{\cal I}_{t+1}(a)=(1/2) \ln[\sigma_a^2(t)\sigma_a^{-2}(t+1)].
	\label{eq:Igauss}
\end{equation}
		
\section{Least Noisy Descattered Reflectivity} 
\label{sec:uncertain}
	
Before underwater optical inspection~\cite{campos2014surface,coiras2009simulation,englot2013three} 
bathymetry (depth mapping) routinely done using Sonar, which penetrates water to great distances. Hence, in relevant applications, the surface topography is roughly available~\cite{campos2014surface}.
At close distance, optical imaging and descattering seeks the spatial distribution of surface albedo
 ${\cal O}=\bigcup_s \rho_s$, to notice sediments, defects in submerged pipes, parasitic colonies in various environments etc.\footnote{In addition, visual data can be integrated to further enhance the topography estimation~\cite{campos2014surface}.} 
Beyond removal of bias by backscatter and attenuation, descattered results need to have low noise variance, so that fine details~\cite{treibitz2009recovery} can be detectable. This is our goal.
	
The {\tt C}\&{\tt L} pose parameters are concatenated into a vector ${\bf v}(t) = [{\bf \phi}_{\tt C}(t),{\bf \phi}_{\tt L}(t)]$. This vector is approximately known during operation, using established localization sensors \cite{maurelli2008particle,paull2014auv,englot2013three}.
Moreover, the water scattering and extinction characteristics are global parameters, that can be measured in-situ. Consequently,  $B_s$ and $E_s$ can be pre-assessed for each ${\bf \phi}_{\tt C}\in\Phi_{\tt C}$,  ${\bf \phi}_{\tt L}\in\Phi_{\tt L}$ and surface patch index $s$.

Using Eq.~(\ref{eq:ImageCreationModel}), descattering based on an image at $t$ is
\begin{equation}
	\hat \rho_s(t) =   [I_s(t) - B_s(t)]/E_s(t).
	\label{eq:hatrrhos}
\end{equation}
Due to noise in $I_s(t)$, the variance of $ \hat \rho_s(t)  $ is:
\begin{equation}
	\sigma_{s}^2(t)=\sigma_{I_s}^2/E_s^2(t) ~.
	\label{eq:hatrrhos_sig}
\end{equation}
Note that $\sigma_{s}^2(t)$ is unknown, since Eqs.~(\ref{eq:ImageCreationModel},\ref{eq:sigmaI}) depend on the unknown $\rho_s$. Nevertheless, it is possible to define an {\em operating point} value for $\rho_s$, by a typical value denoted ${\bar\rho}$. The reason is that, per application, the typical albedos encountered are familiar: typical soil in the known region, anti-corrosive paints in known familiar bridge support etc. The value of ${\bar\rho}$ is rough, but provides a guideline. Consequently
\begin{equation}
	\sigma_{I_s}^2\approx {\bar \sigma}_{I_s}^2\equiv {\bar\rho} E_s(t) + B_s(t)+\sigma_{\text{RN}}^2,
	\label{eq:barIosig}
\end{equation}
\begin{equation}
\begin{gathered}
	\sigma_{s}^2(t) \approx
	\frac{{\bar\rho}_s E_s(t) + B_s(t)+\sigma_{\text{RN}}^2}{E_s^2(t)}
	\equiv  ~1/Q_s(t).
	\label{eq:barrhosig}
\end{gathered}
\end{equation}
Here the defined $Q_s(t)$ is a local quality measure, pre-calculated
$\forall s,{\bf v}$.

Using Eqs.~(\ref{eq:CamProjection},\ref{eq:hatrrhos},\ref{eq:barrhosig}), in the frame of {\tt C}:
\begin{equation}
\label{eq:RhoAndSigImages}
\hat{\rho}_s(t) \rightarrow \hat{\rho}({\bf x},t),~ \sigma^2_s(t)\rightarrow \sigma^2({\bf x},t)
\end{equation}
Components $\hat{\rho}({\bf x},t)$ and $\sigma({\bf x},t)$ are calculated directly using Eqs.~(\ref{eq:hatrrhos},\ref{eq:barrhosig}) and $ E({\bf x},t)$, $ B({\bf x},t) $ and  $ I({\bf x},t) $.

\subsection*{Muti-frame Most-Likely Descattering} 
\label{sec:fuse}	
As described in Sec.~\ref{sec:nbv}, by discrete time $t$, the system has already accumulated
image data $\{I_s(t')\} _{t'=0}^t$. The measurements
have independent noise. Hence, the joint likelihood \mbox{$L_s(t)\equiv L[\{I_s(t')]\}_{t'=0}^t]$} of the data 
is equivalent to the product of probability densities $\forall t'$. Consequently, the log-likelihood is
\begin{equation}
	\tilde L_s(t)=\ln L_s(t)\simeq
	\sum_{t'=0}^t
	\frac{[I_s(t') - B_s(t') -\rho_s E_s(t')]^2 }{{\bar \sigma}_{I_s}^2(t')}.
	\label{eq:logL}
\end{equation}
Differentiating Eq.~(\ref{eq:logL}) with respect to $\rho_s$, the maximum likelihood (ML)
estimator of the descattered $\rho_s$, using all accumulated data is
\begin{equation}
\begin{gathered}
	\hat{\rho}_{s}^{\rm ML}(t)
	=\frac{\sum_{t'=0}^t\hat{\rho_s}(t')[\sigma_s(t')]^{-2}}{\sum_{t'=0}^t[\sigma_s(t')]^{-2}},
\end{gathered}
\label{eq:estimator}
\end{equation}
where $\hat{\rho_s}(t'),\sigma_s(t')$ are derived in Eqs.~(\ref{eq:hatrrhos},\ref{eq:barrhosig}).
The variance of this estimator is
\begin{equation}
	[\sigma_s^{\rm ML}(t)]^2 =
	\left\{
	\sum_{t'=0}^t[\sigma_s(t')]^{-2}
	\right\}^{-1}
	\;\;.
	\label{eq:estimatorSTD}
\end{equation}
From  Eqs.~(\ref{eq:barrhosig},\ref{eq:estimatorSTD}), the quality of the ML descattered 
reflectivity is
\begin{equation}
	\mathcal{Q}^{\rm ML}_{s}(t)\equiv [\sigma_s^{\rm ML}(t)]^{-2}=
	\underbrace{\left[\sum_{t'=0}^{t-1}Q_{s}(t')\right]}_{{\cal Q}^{\rm ML}_s(t-1)} + Q_{s}(t)
	\label{eq:sumQ}
\end{equation}
Eq.~(\ref{eq:sumQ}) expresses how the variance of $\rho^{\rm ML}_s$ can be updated using new data.

\section{Next Best Underwater View} 
\label{sec:NBUV}

After time $t$, the next view ${\bf v}(t+1)$ yields information gain ${\cal I}_{(t+1)}(\mathcal{O})$. Let  $\boldsymbol{\mathcal{V}}$ be the set of all possible (or permissible) camera-lighting poses for time $t+1$. The next underwater view and lighting poses are selected from $\boldsymbol{\mathcal{V}}$, to maximize the information gain measure ${\cal I}_{t+1}(\mathcal{O})$,
\begin{equation}
\hat{\bf v}(t+1) = \underset{{\bf v}\in\boldsymbol{\mathcal{V}}}
{\text{arg~max}} ~{\cal I}_{t+1}(\mathcal{O}).
\end{equation}
We now derive ${\cal I}_{t+1}(\mathcal{O})$ in our case.
Information is an additive quantity for independent measurements. Hence, information gained by enhanced estimation of $ \rho_s $ over $ N_s $ independent surface patches is
\begin{equation}
	{\cal I}_{t+1}(\mathcal{O}) =
	\sum_{s=1}^{N_s} {\cal I}_{t+1}(\hat{\rho}^{\rm ML}_s),
	\label{eq:IGtall}
\end{equation}
From Eq.~(\ref{eq:Igauss}),
\begin{equation}
	{\cal I}_{t+1}(\hat{\rho}^{\rm ML}_s)=
	\frac{1}{2}\ln\left([\sigma_s^{\rm ML}(t)]^2[\sigma_s^{\rm ML}(t+1)]^{-2}\right).
	\label{eq:IGt1}
\end{equation}
From Eqs.~(\ref{eq:sumQ},\ref{eq:IGt1}),
\begin{equation}
	\begin{gathered}
		{\cal I}_{t+1}(\hat{\rho}^{\rm ML}_s) = \frac{1}{2}
		\ln
		  \left[
		     \frac{\sum_{t'=0}^{t+1} Q_{s}(t') }
		          {\sum_{t'=0}^t Q_{s}(t')}
		  \right]=~~~~~~~~~~~~~~~~~~\\
	    ~~~~~~~~~~~~~~~~~~~~~~~
		=\frac{1}{2}\ln\left[1 + \frac{Q_{s}(t+1)}{\mathcal{Q}^{\rm ML}_{s}(t)}\right] .
	\end{gathered}
	\label{eq:IGupdate}
\end{equation}

\section{Path Planning}
\label{sec:pathplan}
Our formalism until now has focused on optimization of the next best view, underwater.
What about next best {\em sequence} of views? Indeed the formalism can be extended to path planning, beyond a single next view. The information gain from  $t$ to $t+1$ is  given by Eqs.~(\ref{eq:IGtall},\ref{eq:IGupdate}). Similarly, the information gain of patch $s$ due to a path from $t_1$ to $t_2$ is 
\begin{equation}
\begin{gathered}
   {\cal I}_{t_1 \rightarrow t_2}(\hat{\rho}^{\rm ML}_s) = \frac{1}{2}
      \ln
     \left[
       \frac{\sum_{t'=0}^{t_2} Q_{s}(t') }
         {\sum_{t'=0}^{t_1} Q_{s}(t')}
       \right]
      =\frac{1}{2}
   \ln
   \left[ 
   \frac{\mathcal{Q}^{\rm ML}_{s}(t_2)}
   {\mathcal{Q}^{\rm ML}_{s}(t_1)}
   \right].
 \end{gathered}
 \label{eq:IGupdatet1t2}
\end{equation}
Thus
\begin{equation}
   {\cal I}_{t_1 \rightarrow t_2}({\cal O})
    =\sum_{s=1}^{N_s}\frac{1}{2}
    \ln
      \left[ 
         \frac{\mathcal{Q}^{\rm ML}_{s}(t_2)}
              {\mathcal{Q}^{\rm ML}_{s}(t_1)}
      \right].
       \label{eq:It1t2}
\end{equation}
A path of {\tt C}\&{\tt L} is 
  ${\cal L} \equiv [{\bf v}(t_1),{\bf v}(t_1+1)\ldots{\bf v}(t_2)]$. Then, in terms of information gain, an optimal path satisfies 
\begin{equation}
\label{eq:optimization}
{\cal L}_{\text{best}}=\underset{\cal L}{\text{argmax}}\left[{\cal I}_{t_1 \rightarrow t_2}({\cal O})\right].
\end{equation} 

\section{Variable Resolution} 
\label{Resolution}

Optimal scanning should strive to provide at least the desired spatial resolution, 
denoted  $R_{\text{min}} [\text{pixels}/ \text{m}^2]$. Over a flat terrain, this requirement is easily met by constraining ${\tt C}$ to be under a specific altitude. Maintaining this altitude maximizes efficiency, since then each image captures the maximal surface area, within this constraint.
In a complex terrain, the trajectory altitude and projected patch resolution vary. This section  describe how calculations are affected. 

At time $t$, patch $s$ is projected to an image segment at resolution $R_{s}(t)$~$[\text{pixels}/ \text{m}^2]$. Define  $\gamma_s(t)\triangleq R_{s}(t)/R_{\text{min}}$. If $\gamma_s(t)<1$, then
patch $s$ appears too small in terms of pixels, which means that camera {\tt C} may be too far from $s$. This may be a problem for patch $s$, but be of benefit to other patches, which are observed better.
We optimize the overall information gain, accounting for all patches. To keep the optimization framework, unconstrained, we took the following step.  When $\gamma_s(t)<1 $, the patch's variance is penalized by:
\begin{equation}
  \sigma^2_{s}(t) ~\leftarrow \sigma^2_{s}(t) \exp(\eta\{ [\gamma_s(t)]^{-1}-1\})
  \label{eq:sigmagamma}
\end{equation}
where $\eta$ is a constant parameter, which we set to 10. We found that this penalty keeps {\tt C} from
distancing from the surface, and provided good results. 

When $\gamma_s(t)>1$, patch $s$ occupies more pixels than the minimum. Pixel redundancy enables digital spatial averaging, which lowers the variance of patch $s$. The image of $s$ occupies a set 
of pixels $U_{s}(t)$. Hence,  when $\gamma_s(t)>1$, spatial averaging sets
\begin{equation}
   \sigma_s^2(t) \equiv 
     \frac{1}{\gamma_s(t)|U_s(t)|} 
      \sum_{ {\bf x} \in U_s(t)} \left[\sigma({\bf x},t)\right]^2,
   \label{eq:sis}
\end{equation}
where $\sigma({\bf x},t)$ is the modeled single-pixel variance (\ref{eq:RhoAndSigImages}).

\begin{figure}[t]
	\def\svgwidth{\columnwidth}
	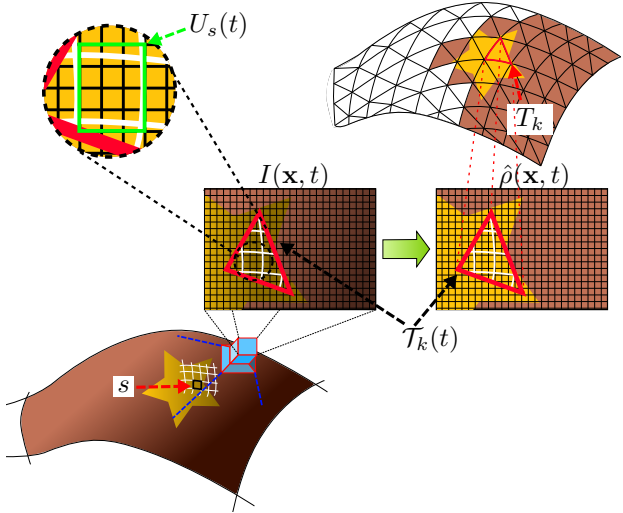
	\caption{\small The geometric transformations between the scene, image plane and texture map.}
	\label{fig:AllTerms}
\end{figure}
\section{Discrete Domain Solution} 
This section discusses how the NBUV method is applied using standard discrete 3D object representations common in computer graphics.

Sonar usually results in a 3D mesh representation of the surface \cite{campos2014surface}. A 3D surface is parameterized by a triangulated mesh ${\cal M}=({\cal F},{\cal E}) $, where ${\cal F}$ are the mesh's faces and $ {\cal E} $ are the edges. To reduce complexity, the surface is divided into  $N_{\rm m}$ non-uniform segments $ \{T_{k}\}^{N_{\rm m}}_{k=1} $, where $ k $ is the index of the segment. Each segment contains several patches $s$ within its area  $ |T_{k}| $. Let $ \lambda_k \triangleq |T_{k}|/|s| $ be the number of patches in the segment. In most cases, it is convenient to use the faces of $ {\cal M} $ as the segments, i.e. $ T_{k}\triangleq {\cal F}[k] $, where ${\cal F}[k] $ is the $ k $'th face. At time $ t $ segment $ T_k $ is projected to a set of pixels $ {\cal T}_k(t) $ as illustrated in Fig.~\ref{fig:Fusing}b-c. For a triangulated mesh representation, the support of $ {\cal T}_k(t) $ on the image plane is a triangle. The size of $ {\cal T}_k(t) $ is $|{\cal T}_k(t)|$.

\begin{figure}[t]
	\def\svgwidth{\columnwidth}
	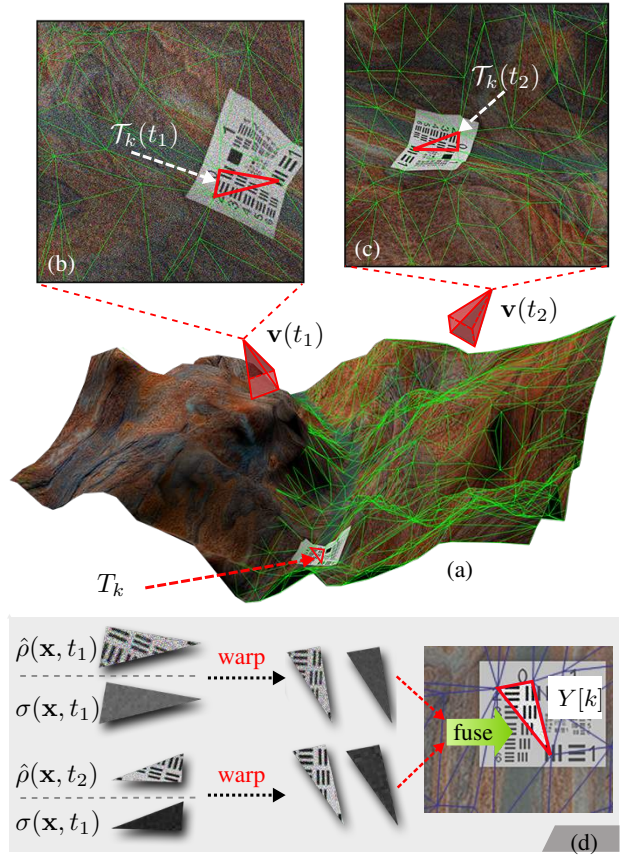
	\caption{\small (a) A surface is imaged from two poses ${\bf v}(t_1)$ and ${\bf v}(t_2)$. Green mesh represents segmentation $ T_{k}$. (b) Image from ${\bf v}(t)$, ${\cal T}_k(t_1)$ is marked in red. $ \gamma_k(t)$ is slightly larger than 1. (c) Image from ${\bf v}(t_2)$, $ \gamma_k(t+1)$ is slightly smaller than 1. (d) Imaged segments are warped and scaled to match $ Y[k] $.  Note how $\sigma_{\tt C}^{(t)}[{\bf x}]$ is scaled according to $ \gamma_k(t) $ for both measurements.}
	\label{fig:Fusing}
\end{figure}

\subsection{Fusing Measurements}
The goal of estimation process of $ \hat{\rho}_s $ is to produce a texture map for $ \cal M $. Each mesh triangle $ {\cal F}[k] $ is linearly mapped onto a corresponding triangle $ Y[k] $ in a  'texture' image, (Fig.~\ref{fig:Fusing}e). If $ T_k $ is imaged once at $ t_0 $, no fusing is required. Then, $ T_k $ is mapped to $ {\cal T}_k(t_0) $ at image $ \hat{\rho}({\bf x},t_0) $ (Fig.~\ref{fig:AllTerms}).
However, if $ T_k $ is imaged more than once, all its imaged segments are fused. Fusing is done as described in sec.~\ref{sec:fuse}. The fusing process weights the image formation noise as well as the scanning resolution. For a detailed description of the fusing process see Appendix.

\subsection{Information Gain Calculation}
Evaluating the information gain over the surface requires storing the uncertainty of all patches. To avoid this the information gain is calculated per segment. For \textit{segments}, the information gain over the entire surface is calculated using:
\begin{equation}
{\cal I}_{t+1}({\cal O}) \approxeq \sum_{k=1}^{N_{\rm m}} \left(\frac{1}{2} \ln \left[ 1+\frac{Q_{k}(t+1)}{{\cal Q}^{\rm ML}_{k}(t)} \right] \cdot\lambda_k \right),
\label{eq:IGbySegments}
\end{equation}
Eq.~(\ref{eq:IGbySegments}) is derived from Eq.~(\ref{eq:IGupdate}), except that now the quality measure is attributed to the segments:
\begin{equation}
Q_{k}(t)^{-1} = \left(\tilde{\sigma}_{k}(t) \cdot \exp(\eta\{ [\gamma_k(t)]^{-1}-1\} \right)^2,
\end{equation}
where $\tilde{\sigma}_{k}(t)$ is the mean uncertainty of pixels belonging to $ {\cal T}_k(t) $ in $ \sigma({\bf x},t)$ (Eq.~\ref{eq:RhoAndSigImages}):
\begin{equation}
 \tilde{\sigma}_k(t)\simeq\frac{1}{|{\cal T}_k(t)|}\sum_{{\bf x} \in {\cal T}_k(t)} \sigma({\bf x},t).
\label{eq:SigmaOfTn}
\end{equation}
As in Eq.~(\ref{eq:sigmagamma}), the uncertainty is factored by $ T_k $'s scale at time $ t $: $ \gamma_k(t)\triangleq R_k(t)/R_\text{min} $, where $ R_k(t) $ is the effective resolution of $ T_k $ at time $ t $.
Finally we calculate the information gain on the entire surface using Eq.~(\ref{eq:IGupdate}):
where $ \mathcal{Q}^{\rm ML}_k(t) = \sum_{t'=0}^t Q_k(t)$ following Eq.~(\ref{eq:sumQ}).
Eq.~(\ref{eq:IGbySegments}) assumes that the variance in noise affecting patches in each segment is small. This assumption improves with a finer parametrization of the surface.

\section{Simulations}
We use the scattering model of \cite{gupta2008controlling}, in a homogeneous medium. This model renders the imaged radiance in Eq.~(\ref{eq:ImageCreationModel}). The surface is Lambertian. The validity of a Lambertian assumption increases underwater \cite{schechner2011spaceborne,zhang2006bidirectional}.
We set $ \sigma_{\text{RN}} = 13.1$[e] and a max well of 24,000 [e] in a perspective camera {\tt C} based on Canon 60D camera data, while {\tt L} is a spot light with no lateral falloff.

\begin{figure}[t]
	\def\svgwidth{\columnwidth}
	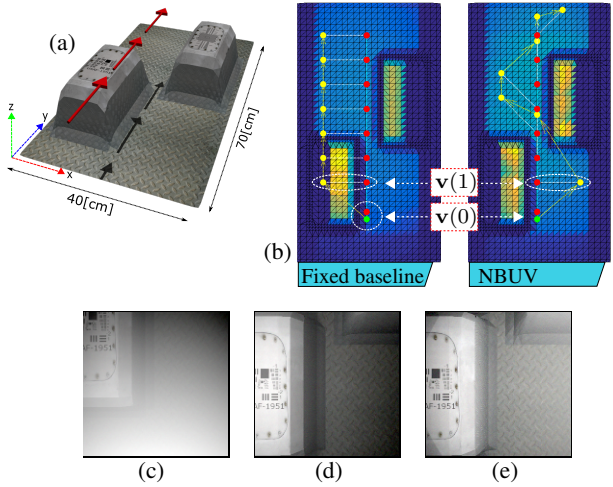
	\caption{\small Simulation. (a) The scanned surface, the camera's scanning trajectory is along the red arrows. (b) Trajectories of fixed-baseline and NBUV. Face color represent  $Q_k(t=8)$. Note how the NBUV avoids casting shadows from the hills to the floor. (c) Image take from $ {\bf v}(0) $, showing significant backscatter due to a small {\tt LC} baseline. (d)-(e) Images from ${\bf v}(1)$ using a fixed-baseline and NBUV methods, respectivly.}
	\label{fig:SimulationResult1}
\end{figure}


Fig.~\ref{fig:SimulationResult1} illustrates a simple scenario.
A straight path 40cm above a surface is set. The medium's parameters are 
$\beta=5 [1/{\rm m}]$, while the anisotropy parameter is $g=0.6$ in the Henyey-Greenstein phase function~\cite{gupta2008controlling,haltrin2002one}. {\tt C}\&{\tt L} start from $ {\bf v}(0) $. The initial {\tt LC} baseline is 2cm. Such a baseline would be fine in a clear medium. Underwater however, this baseline results in significant backscatter (Fig.~\ref{fig:SimulationResult1}c). Hence, a 12cm baseline is used. The scan consists of 8 views, where $ {\bf\phi}_{\tt C}(t) $ is spread uniformly across the path.

In a traditional path, $\vec{\tt LC}=12\hat{x} $, while {\tt L} is points to the center of {\tt C}'s field of view on the surface. To the best of our knowledge, prior dehazing methods ignore SNR variability in image sequences. Thus, simple averaging is used for $\hat{\rho_s}$ in a traditional process. When we ran our optimization, $ {\bf \phi}_{\tt L}(t+1) $ is selected out of a set of 32 possible locations with different radii around $ {\bf\phi}_{\tt C}(t) $. In addition, in each location there are 9 orientations of $ {\bf\phi}_{\tt L} $, facing nadir, $\pm 10^{\circ}$ or 
$\pm 20^{\circ} $ off nadir to each lateral-direction. Then, NBUV is chosen out of a total number of $|\boldsymbol{\mathcal{V}}(t)|= 288$ by exhaustive search.

Fig.~\ref{fig:SimulationResult1}b shows the two {\tt C}\&{\tt L}  trajectory. Looking at the simple geometry under scan, it is clear that the illumination should be from the opposite side of the hills, when the camera passes above them. This is evident in $ {\bf v}(1) $ (Fig.~\ref{fig:SimulationResult1}d-e), where the view chosen by the NBUV method is lit better than the fixed baseline one.


Fig.~\ref{fig:SimulationResult2}. illustrates how NBUV is used to determine an optimal scanning path. A cube having 28cm edge length is placed on a flat surface in scattering medium 
 ($\beta=2.5 [1/{\rm m}],~g=0.6$). A trivial scanning path 84cm above the surface is set over the scene. The path consists of 6 uniformly distributed views. To avoid backscatter the baseline is $ \vec{\tt LC}=\text{34cm} $. Let $ {\cal L} = \{ {\bf v}(1) , {\bf v}(2)..{\bf v}(6)\}$ denote the scanning path. Optimization was initialized with the trivial path. Optimization was performed using 20 iterations of Matlab's direct search function \cite{kolda2006generating} over the $ \cal L $'s 60 degrees of freedom.\footnote{Rotation about the $ Z $ axis was excluded for both $ \tt L $ and $ \tt C $.}

In the initial trivial scan, the left and right faces (see Fig.~\ref{fig:SimulationResult2} red arrow) are occluded as $ \tt C $ passes over the cube. In addition, shadow from the cube heavily degrades the left side of the surface. Applying our method, $ \tt C $ and $ \tt L $ are moved to cover the occluded regions (Fig.~\ref{fig:SimulationResult2} views 3-4). Note that the front and back faces of the cube (blue arrow in Fig.~\ref{fig:SimulationResult2}) are scanned in better resolution. This is owed to views 2 and 5 (Fig.~\ref{fig:SimulationResult2} ) that changed perspective to scan these faces. In total, our method reduced the total estimation uncertainty by ~30\% over the trivial setup. 

\begin{figure}[t]
\def\svgwidth{\columnwidth}
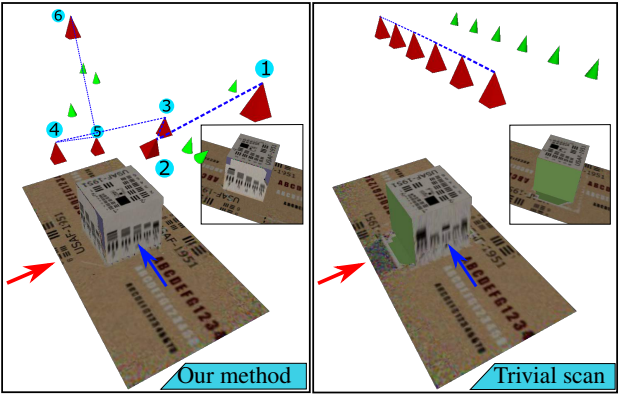
	\caption{\small Path planning. Red cones - $ \tt C $. Green cones - $ \tt L $. Top, scanning result. (a) Image from ${\bf v}(2)$ in trivial scan. (b) Image from ${\bf v}(3)$ in trivial scan. Notice the shadowed surface to the left. Left and right faces are occluded in trivial scan (Red arrow). A total of ~30\% improvement in estimation uncertainty (e.g. red and blue arrows).}
	\label{fig:SimulationResult2}
\end{figure}
\section{Experiment}
\begin{figure}[t]
	\includegraphics[width=\columnwidth]{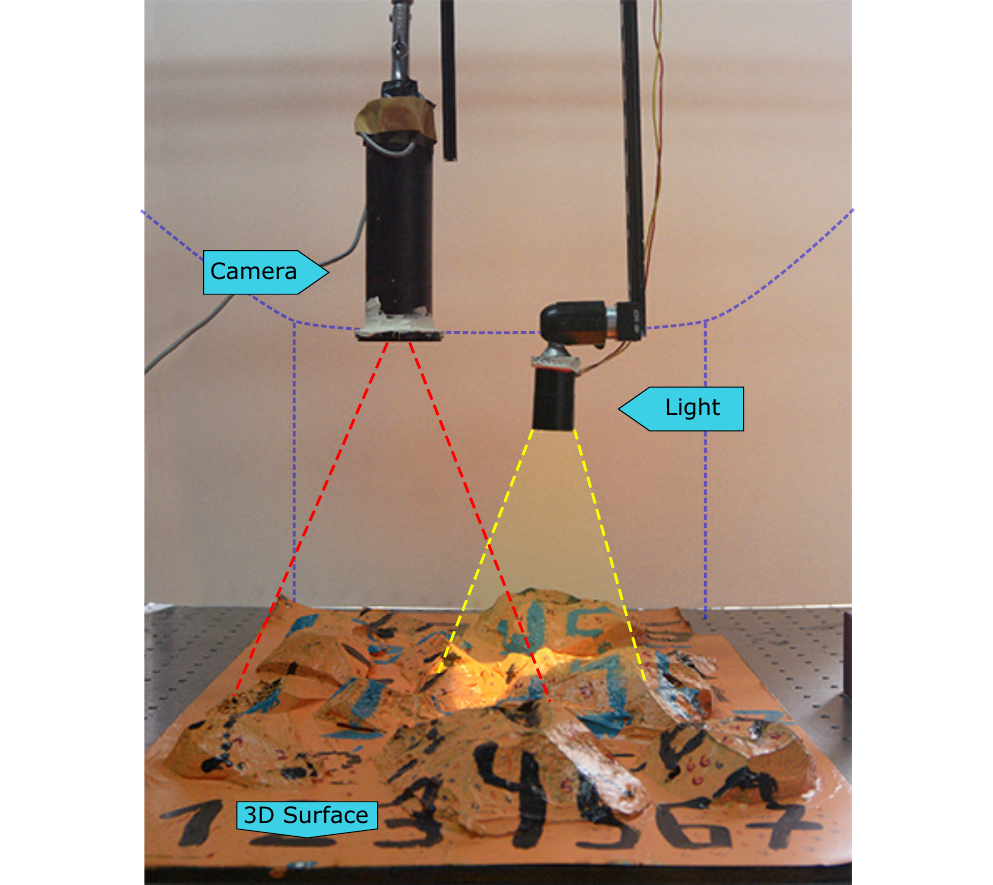}
	\caption{\small Experiment setup. A camera in a watertight housing images a surface submerged in a small water tank. An LED in a cylinder creates a light cone.}
	\label{fig:ExperimentSetup}
\end{figure}
We built a model having an arbitrary non-trivial topography submerged in water (Fig.\ref{fig:ExperimentSetup}). Emulating a sonar scan, the surface model was pre-scanned using a depth camera in a clear medium to produce a 3D mesh. Mixing milk in the water then produced single scattering conditions that fit our image formation model~\cite{narasimhan2006acquiring}. A machine vision camera was submerged in a watertight housing. The submerged surface was illuminated using a Mouser Electronics, 'Warm White' 3000K LED. The intrinsic parameters of the camera and the illumination angle of the LED were both calibrated underwater to account for the water's refractive index. A robotic 2D plotter was used to move the light and camera to their approximate locations in space.

The medium's parameters ($ \beta $ and $ g $) were extracted in-situ. The camera and LED light were placed in a known state above a flat white sheet ($\rho_s\approxeq1$).  Optimization was used to extract $ \beta $ and $ g $, namely:
\begin{equation}
g,\beta = \underset{g,\beta}{\text{argmin}}\|E({\bf x},t) + B({\bf x},t)  - I({\bf x},t) \|^2_F.
\label{eq:WaterOptimization}
\end{equation}
\begin{figure}[t]
	\def\svgwidth{\columnwidth}
	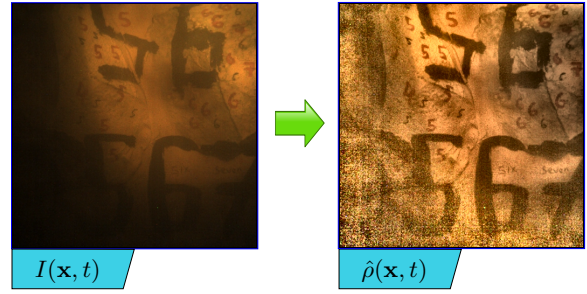
	\caption{\small Albedo extraction. Left: Scanning image from experiment. Right: Recovered albedo image, extracted using Eq.~(\ref{eq:AlbedoExtraction}).}
	\label{fig:AlbedoRecovery}
\end{figure}
\subsection{Numerical Conditioning}
Using Eq.~(\ref{eq:hatrrhos}) directly to recover $ \hat{\rho}({\bf x},t) $ may lead to unstable results. Shadowed image regions where $ \cal M $ poorly approximates the topography, may cause $E({\bf x},t)$ to be lower than desired. Lower $E({\bf x},t)$ stem from deviation from the imaging model, e.g. the scattering order in. Since $E({\bf x},t) $ is the denominator of Eq.~(\ref{eq:hatrrhos}), $ \hat{\rho}({\bf x},t) $ may exceed 1 in these regions.
Therefore, $E({\bf x},t)$ is stabilized by:
\begin{equation}
\label{eq:AlbedoExtraction}
\begin{gathered}
\tilde{E}({\bf x},t) = \\
= \left(\left[E({\bf x},t)\ast h_E\right] (1-w) + \left[I({\bf x},t)\ast h_I\right]w\right)\ast h_T,
\end{gathered}
\end{equation}
where $ h_E $, $ h_I $, and $ h_{\rm T} $ are Gaussian convolution kernels, and $ w $ is an alpha mask. $ w{(\bf x)}=1$ whenever $\hat{\rho}({\bf x},t)>1$.
An example of the recovery result can be seen in Fig.~(\ref{fig:AlbedoRecovery}).
\begin{figure*}[t]
	\includegraphics[width=\textwidth]{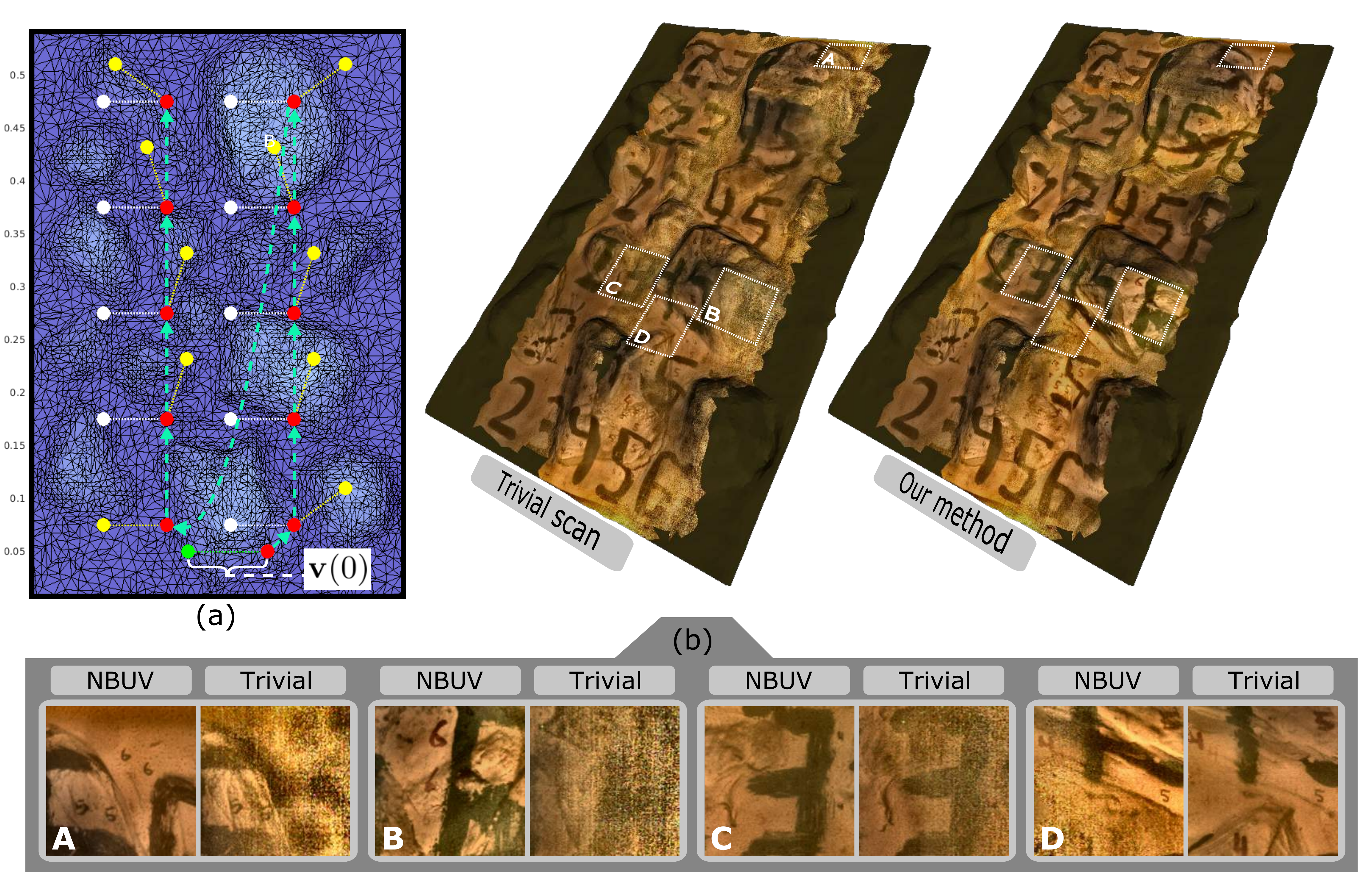}
	\caption{\small Experiment results. (a) Scanning path and light locations, Red - camera locations, Yellow - NBUV lights, White - Fixed baseline lights, Green - Initial light position. (b) Fixed-baseline scanning result. (c) NBUV scanning result. (d) Close up comparison. A-C show improved estimation quality while D shows a patch where fixed baseline produced better estimation.}
	\label{fig:ExperimentResults}
\end{figure*}
\subsection{Results}
In a standard fixed-baseline scheme, the camera and light are initialized in a certain state above the model (Figure \ref{fig:ExperimentResults} a). Ten uniformly-spaced camera locations are used. Due to mechanical limitations, we allow the camera and light to move only horizontally. The elevation of the camera and light is set to 20cm above the object, and both are facing directly down.

In the fixed-baseline configuration the light is placed in a fixed distance from the camera so as to avoid significant backscatter. In our NBUV setup, per $ t $ we allow the light to be placed in 40 fixed states around the location of the camera ${\bf \phi}_{\tt C}(t)$. We use our method to simulate the expected IG from each optional camera-light state ${\bf v}(t)$, which in this case is chosen out of $|\boldsymbol{\mathcal{V}}(t)|=40~~ \forall t\in[1..10]$. Both configurations are initialized at $ {\bf v}(0) $.

A total of 11 views of the surface are taken in both fixed-baseline and NBUV configurations.
The recovered albedo images were mapped to the 3D mesh of the surface (Fig.~\ref{fig:ExperimentResults}b-c). Our method provides a better overall surface estimation over the fixed-baseline configuration. For the particular surface we used, that meant lighting dark spots on the left and center of the surface, where the shadows were cast by the fixed light due to the topography. Fig.~\ref{fig:ExperimentResults}d A-C show areas where the expected estimation noise is significantly lower, when using NBUV. The total estimation uncertainty over the entire surface is lower. However, not all surface patches benefit Fig.~\ref{fig:ExperimentResults}d D.

\section{Conclusions}

The paper defines the next-best-view task, as well as optimized path planning, by taking into account scattering effects. The method optimizes viewpoints so the descattered albedo is least noisy, allowing resolution of fine details. It generalizes dehazing to scanning multiview platforms. We believe this approach can make drone imaging flights and underwater robotic imaging significantly more efficient when operating in poor visibility conditions due to scattering effects. Further work can use more comprehensive scattering models, image statistics priors and path-length penalties. Moreover, the principle we proposed can benefit from optimization algorithms that are more efficient, as the number of degrees of freedom increases.  The work here can be generalized to multiple agents (cameras) cooperatively scanning the scene.

	\section*{Appendix A: Discrete Domain Fusing}
	\label{App:A}	
	The estimation process produces a texture map for $ \cal M $. Each mesh triangle $ {\cal F}[k] $ is linearly mapped onto a corresponding triangle $ Y[k] $ in a 'texture' image, (Fig.~\ref{fig:Fusing}e). If $ T_k $ is imaged once at $ t_i $, then $ Y[k]\equiv{\cal T}_k(t_i) $ on image $\hat{\rho}({\bf x},t_0)$. On the other hand, if $ T_k $ is imaged more than once, e.g.~at $ t_j:j\in\{1,2,5\} $, all imaged segments (triangles) $ {\cal T}_k(t_j)$ in corresponding images $\hat{\rho}({\bf x},t_j)$ are fused.
	Each segment $ T_k $ may be imaged with different resolutions.
	$Y[k]$ is allocated in a new image at resolution $R_\text{min}$.
	The supports of $ {\cal T}_k(t_j) $ in both $\hat{\rho}({\bf x},t_j)$ and $ \sigma({\bf x},t_j)$ are scaled by $ \gamma_k(t_j)^{-1} $ and transformed to align with $Y[k]$ (see Fig.~\ref{fig:Fusing}d). Once aligned, $\hat{\rho}({\bf x},t_j):{\bf x}\in {\cal T}_k(t_j) $ are fused using Eq.~(\ref{eq:estimator}). $ \sigma({\bf x},t_j)$ used in Eq.~(\ref{eq:estimator}) is factored according to $ \gamma_k(t_j) $ as described in sec. \ref{Resolution}.

{\small

\begin{thebibliography}{10}\itemsep=-1pt
	
	\bibitem{ahmed1989entropy}
	N.~A. Ahmed and D.~Gokhale.
	\newblock Entropy expressions and their estimators for multivariate
	distributions.
	\newblock {\em IEEE Trans. on IT}, 35:688--692, 1989.
	
	\bibitem{anguelov2010google}
	D.~Anguelov, C.~Dulong, D.~Filip, C.~Frueh, S.~Lafon, R.~Lyon, A.~Ogale,
	L.~Vincent, and J.~Weaver.
	\newblock Google street view: Capturing the world at street level.
	\newblock {\em Computer}, (6):32--38, 2010.
	
	\bibitem{campos2014surface}
	R.~Campos, R.~Garcia, P.~Alliez, and M.~Yvinec.
	\newblock A surface reconstruction method for in-detail underwater 3d optical
	mapping.
	\newblock {\em Int. J. Robot. Res.}, vol. 34, pp. 64–89, 2014.
	
	\bibitem{chen2005vision}
	S.~Chen and Y.~Li.
	\newblock Vision sensor planning for 3-d model acquisition.
	\newblock {\em Systems, Man, and Cybernetics, Part B: Cybernetics, IEEE
		Trans. on}, 35(5):894--904, 2005.
	
	\bibitem{coiras2009simulation}
	E.~Coiras and J.~Groen.
	\newblock {\em Simulation and 3D reconstruction of side-looking sonar images}.
	\newblock INTECH Open Access Publisher, 2009.
	
	\bibitem{cowan1992automatic}
	C.~K. Cowan, B.~Modayur, and J.~L. DeCurtins.
	\newblock Automatic light-source placement for detecting object features.
	\newblock In {\em Applications in Optical Science and Engineering}, pages
	397--408. SPIE, 1992.
	
	\bibitem{dalgleish2013extended}
	F.~R. Dalgleish, A.~K. Vuorenkoski, and B.~Ouyang.
	\newblock Extended-range undersea laser imaging: Current research status and a
	glimpse at future technologies.
	\newblock {\em Marine Technology Society Journal}, 47(5):128--147, 2013.
	
	\bibitem{fattal2008single}
	R.~Fattal.
	\newblock Single image dehazing.
	\newblock In {\em ACM TOG}, volume~27, page~72. ACM,
	2008.
	
	\bibitem{gkioulekas2015micron}
	I.~Gkioulekas, A.~Levin, F.~Durand, and T.~Zickler.
	\newblock Micron-scale light transport decomposition using interferometry.
	\newblock {\em ACM TOG}, 34(4):37, 2015.
	
	\bibitem{gupta2008controlling}
	M.~Gupta, S.~G. Narasimhan, and Y.~Y. Schechner.
	\newblock On controlling light transport in poor visibility environments.
	\newblock In {\em Computer Vision and Pattern Recognition, 2008. CVPR 2008.
		IEEE Conference on}, pages 1--8. IEEE, 2008.
	
	\bibitem{haltrin2002one}
	V.~I. Haltrin.
	\newblock One-parameter two-term henyey-greenstein phase function for light
	scattering in seawater.
	\newblock {\em Applied Optics}, 41(6):1022--1028, 2002.
	
	\bibitem{he2011single}
	K.~He, J.~Sun, and X.~Tang.
	\newblock Single image haze removal using dark channel prior.
	\newblock {\em IEEE T.PAMI}, 33(12):2341--2353, 2011.
	
	\bibitem{heide2014imaging}
	F.~Heide, L.~Xiao, A.~Kolb, M.~B. Hullin, and W.~Heidrich.
	\newblock Imaging in scattering media using correlation image sensors and
	sparse convolutional coding.
	\newblock {\em Optics express}, 22(21):26338--26350, 2014.
	
	\bibitem{jaffe1990computer}
	J.~S. Jaffe.
	\newblock Computer modeling and the design of optimal underwater imaging
	systems.
	\newblock {\em EEEJ. Oceanic Eng.}, 15(2):101--111, 1990.
	
	\bibitem{jaffe2007multi}
	J.~S. Jaffe.
	\newblock Multi autonomous underwater vehicle optical imaging for extended
	performance.
	\newblock In {\em OCEANS 2007-Europe}, pages 1--4. IEEE, 2007.
	
	\bibitem{kolda2006generating}
	T.~G. Kolda, R.~M. Lewis, and V.~Torczon.
	\newblock A generating set direct search augmented lagrangian algorithm for
	optimization with a combination of general and linear constraints.
	\newblock {\em Sandia National Laboratories}, 2006.
	
	\bibitem{maurelli2008particle}
	F.~Maurelli, S.~Krupi{\'n}ski, Y.~Petillot, and J.~Salvi.
	\newblock A particle filter approach for AUV localization.
	\newblock In {\em OCEANS 2008}, pages 1--7. IEEE, 2008.
	
	\bibitem{narasimhan2006acquiring}
	S.~G. Narasimhan, M.~Gupta, C.~Donner, R.~Ramamoorthi, S.~K. Nayar, and H.~W.
	Jensen.
	\newblock Acquiring scattering properties of participating media by dilution.
	\newblock {\em ACM TOG}, 25(3):1003--1012, 2006.
	
	\bibitem{norwich1993information}
	K.~H. Norwich.
	\newblock {\em Information, sensation, and perception}.
	\newblock Academic Press San Diego, 1993.
	
	\bibitem{o2012primal}
	M.~O'Toole, R.~Raskar, and K.~N. Kutulakos.
	\newblock Primal-dual coding to probe light transport.
	\newblock {\em ACM TOG}, 31(4):39, 2012.
	
	\bibitem{paull2014auv}
	L.~Paull, S.~Saeedi, M.~Seto, and H.~Li.
	\newblock AUV navigation and localization: A review.
	\newblock {\em EEEJ. Oceanic Eng.}, 39(1):131--149, 2014.
	
	\bibitem{pito1999solution}
	R.~Pito.
	\newblock A solution to the next best view problem for automated surface
	acquisition.
	\newblock {\em Pattern Analysis and Machine Intelligence, IEEE Transactions
		on}, 21(10):1016--1030, 1999.
	
	\bibitem{ratner2007illumination}
	N.~Ratner and Y.~Y. Schechner.
	\newblock Illumination multiplexing within fundamental limits.
	\newblock In {\em Computer Vision and Pattern Recognition, 2007. CVPR'07. IEEE
		Conference on}, pages 1--8. IEEE, 2007.
	
	\bibitem{roman2007self}
	C.~Roman and H.~Singh.
	\newblock A self-consistent bathymetric mapping algorithm.
	\newblock {\em J. of Field Robotics}, 24(1-2):23--50, 2007.
	
	\bibitem{sakane1991automatic}
	S.~Sakane and T.~Sato.
	\newblock Automatic planning of light source and camera placement for an active
	photometric stereo system.
	\newblock In {\em Robotics and Automation, 1991. Proceedings., 1991 IEEE
		International Conference on}, pages 1080--1087. IEEE, 1991.
	
	\bibitem{schechner2011spaceborne}
	Y.~Y. Schechner, D.~J. Diner, and J.~V. Martonchik.
	\newblock Spaceborne underwater imaging.
	\newblock In {\em Proc. IEEE ICCP}, pages 1--8. 2011.
	
	\bibitem{treibitz2009active}
	T.~Treibitz and Y.~Y. Schechner.
	\newblock Active polarization descattering.
	\newblock {\em Pattern Analysis and Machine Intelligence, IEEE Transactions
		on}, 31(3):385--399, 2009.
	
		\bibitem{treibitz2009recovery}
		T.~Treibitz and Y.~Y. Schechner.
		\newblock Recovery limits in pointwise degradation.
		\newblock In {\em Computational Photography (ICCP), 2009 IEEE Inter.
			Conf. on}, pages 1--8. IEEE, 2009.
		
	\bibitem{treibitz2012turbid}
	T.~Treibitz and Y.~Y. Schechner.
	\newblock Turbid scene enhancement using multi-directional illumination fusion.
	\newblock {\em Image Processing, IEEE Transactions on}, 21(11):4662--4667,
	2012.
	
	\bibitem{wenhardt2006information}
	S.~Wenhardt, B.~Deutsch, J.~Hornegger, H.~Niemann, and J.~Denzler.
	\newblock An information theoretic approach for next best view planning in 3-d
	reconstruction.
	\newblock In {\em Pattern Recognition, 2006. ICPR 2006. 18th International
		Conference on}, volume~1, pages 103--106. IEEE, 2006.
	
	\bibitem{zhang2006bidirectional}
	H.~Zhang and K.~J. Voss.
	\newblock Bidirectional reflectance study on dry, wet, and submerged
	particulate layers: effects of pore liquid refractive index and translucent
	particle concentrations.
	\newblock {\em Applied optics}, 45(34):8753--8763, 2006.

	\bibitem{englot2013three}
	B.~Englot and F.~S. Hover.
	\newblock Three-dimensional coverage planning for an underwater inspection
	robot.
	\newblock {\em The International Journal of Robotics Research},
	32(9-10):1048--1073, 2013.


\end{thebibliography}

\end{document}